\pdfoutput=1

\documentclass[11pt]{article}

\usepackage[]{emnlp2021}

\usepackage{times}
\usepackage{latexsym}

\usepackage[T1]{fontenc}

\usepackage[utf8]{inputenc}

\usepackage{microtype}
\usepackage{comment}
\usepackage{graphicx}
\graphicspath{ {./images/} }

\usepackage{booktabs}
\newcommand{\mc}[3]{\multicolumn{#1}{#2}{#3}}

\title{DFKI-MLT at WMT-SLT22\\ Spatio-temporal Sign Language Representation and Translation}

\author{Yasser Hamidullah \and Josef van Genabith \and Cristina España-Bonet \\ { \tt \{yasser.hamidullah,Josef.van{\_}Genabith,cristinae\}@dfki.de} \\German Research Center for Artificial Intelligence (DFKI)\\
Saarland Informatics Campus, Saarbrücken, Germany
}

\begin{document}
\maketitle
\begin{abstract}
This paper describes the DFKI-MLT submission to the WMT-SLT 2022 sign language translation (SLT) task from Swiss German Sign Language (video) into German (text).
State-of-the-art techniques for SLT use a generic seq2seq architecture with customized input embeddings. Instead of word embeddings as used in textual machine translation, SLT systems use features extracted from video frames. Standard approaches often do not benefit from temporal features. In our participation, we present a system that learns spatio-temporal feature representations and translation in a single model, resulting in a real end-to-end architecture expected to better generalize to new data sets. Our best system achieved $5\pm1$ BLEU points on the development set, but the performance on the test dropped to $0.11\pm0.06$ BLEU points.
\end{abstract}

\section{Introduction}
\label{s:intro}
Text-to-text machine translation (MT) is achieving a great success with even (close to) human performance for some language pairs and domains~\cite{wmt21}. However, the situation in sign language translation (SLT) is much different.
One important reason is that the SLT is a low-resource scenario where one does not have the same amount of data as in high-resourced text-to-text to achieve a similar level of performance. A more specific reason is that SLT involves two modalities, text and video.
Various problems arise when dealing with these modalities. Besides data scarcity, the lack of temporal boundaries in the input videos is a challenge. To overcome the lack of temporal boundaries, the most common solution tends to ignore or not benefit from temporal features. This approach relies on the Transformer \cite{Vaswani:2017}
capabilities to learn sequence-to-sequence tasks. The state-of-the-art SLT technique \citep{camgoz2020} is practically a normal Transformer but uses a custom embedding layer for 2D features extracted from video frames. In this approach, training a SLT system requires a pre-extraction step to convert frame features to vectors and train a Transformer separately to translate the vectors into spoken language. This type of approach has been widely used on a very specific dataset, the weather forecast corpus PHOENIX14T~\cite{CamgozEtAl:2018}, where researchers reported a relatively good performance in terms of BLEU ($\sim$20) \citep{camgoz2020,Yuecong2021}. 

Despite its good performance on a specific dataset, there is the doubt whether such type of architecture generalizes to new data sets.
In order to build a more general technique, we focus on fundamental SLT problems such as the design, implementation and evaluation of a fully end-to-end model and representation learning for sign language videos. 
Having a fully end-to-end model facilitates the task of data collection and diminishes the need for annotation (e.g. in terms of sign language glosses), which is necessary to build larger and richer datasets. 
It also allows training video embeddings fully optimized for the translation task.
Text translation is one of the most mature areas in natural language processing, and therefore we focus here on the sign language representation part of the architecture and use an in-house state-of-the-art Transformer for text generation.

This paper reports our approach for end-to-end SLT used for the WMT-SLT translation shared task from Swiss German Sign Language into German.  In the next sections we introduce our approach (Section~\ref{s:approach}), experimental setup (Section~\ref{s:experiments}), results (Section~\ref{s:results}) and conclusions \& perspectives (Section~\ref{s:conclusions}).

\begin{figure*}[!ht]
  \centering
  \includegraphics[scale=0.5]{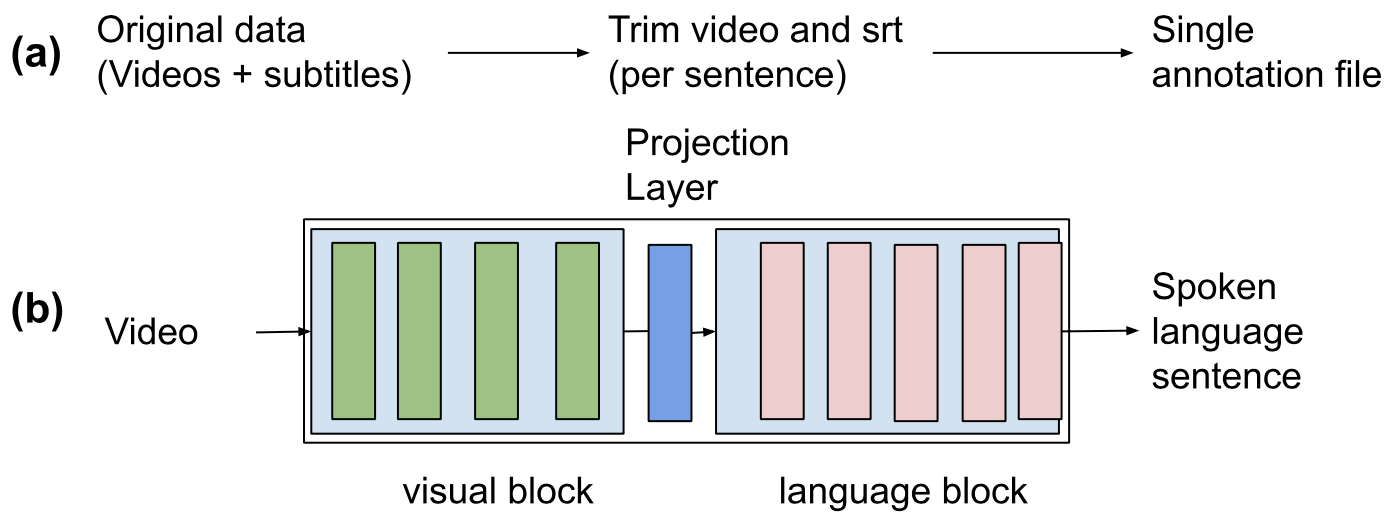}
  \caption{End-to-end architecture for sign language translation.}
  \label{fig:archm}
\end{figure*}

\section{Our Approach}
\label{s:approach}
The main idea of our approach is to learn feature representation and translation in a single model, and be able to train them together. Figure \ref{fig:archm}(a) sketches our general pre-processing pipeline and Figure \ref{fig:archm}(b) the architecture. The system architecture consists of two connected blocks: the first block made of CNNs is intended for vision and the second one is for language which is Transformer architecture.

Both the CNNs used for the video representations and the Transformers used for the text representations come with large numbers of parameters. As we are operating in a low resource scenario, besides the combination of the two networks, we experiment to find the best trade-off between data size and number of parameters. 

\subsection{Visual feature representation}
Our goal is to build a sentence embedding-like model for the visual sign language encoder, as a word/sign level-like representation is limited by the lack of temporal boundaries in videos. We hypothesize that a sentence embedding will still contain and distinguish all the information given by individual signs.

In the shared task, we use ResNet3D \cite{Kensho17} as our spatio-temporal visual feature representation block. We prefer it instead of a normal 2D with temporal convolutions \cite{wangtsn17} to develop a fully end-to-end trainable model. There are many available architectures in the literature, but ResNet is unique in providing different models with different scales.
This gives us the possibility to experiment with various sizes 
which help us to weight the importance of each of the vision and text blocks in our trade-off experiments. 

Our visual encoding in the submitted system is composed by the original 3D ResNet10 with output conversion. The conversion  creates a sequence of vectors from the single output vector to adapt to the transformer encoder input. We define the \textbf{SWM} parameter (Sentence to Words Mapping), which is the number of splits from the output vector.
This output is projected through a linear layer which is connected directly to the language block.
We experiment with 3D ResNet10, 3D ResNet34 and 3D ResNet50 and show the comparative results in Section~\ref{s:experiments}.

\subsection{Language representation}
The language block is a normal language Transformer. 
Its training end-to-end with the visual model can constrain the visual model and force it to take into account the language representation to build the visual embedding. This should result in more specific visual representations for sign language which has not yet been explored extensively in SLT. For this shared task, we use the Transformer for the language block with parameters shown in Table~\ref{t:protoc}. This choice is motivated by \citet{camgoz2020} which improved their previous results with LSTMs and GRUs \cite{CamgozEtAl:2018} by more than 10 BLEU points. Furthermore, the Transformer makes the visual and language fusion more intuitive and easier for SLT, because it can process the whole sentence at the same time.

\begin{table*}[t]
\centering
 \resizebox{\textwidth}{!}{%
\begin{tabular}{|l|l|l|} 
\hline
\textbf{Parameter}                 & \textbf{Value}                                                               & \textbf{Comments}                                                                                                     \\ 
\hline
Training corpus         & FN+SRF                                                                       & Remove sentences with >50 tokens                                                                     \\ 
\hline
Batch size                 & 10                                                                           & Using few workers (<=5) on a single GPU                                                                                  \\ 
\hline
End of training criteria   & PPL                                                                          & \begin{tabular}[c]{@{}l@{}} Stop after 14 epochs without improvements\end{tabular}     \\ 
\hline
Language model             & Transformer "base"                                                            & The number of encoder/decoder layers is 3 instead of 6                                                                   \\ 
\hline
Visual model               & \begin{tabular}[c]{@{}l@{}}3D ResNet \\(outsize= 2048, depth=50)\end{tabular} & \begin{tabular}[c]{@{}l@{}}Additional custom module that converts the  \\ output size to our Sentence to Words Mapping (SWM)  \end{tabular}                                                           \\ 
\hline
SWM                        & 32                                                                           & \begin{tabular}[c]{@{}l@{}}Numbof the splits.\end{tabular}  \\ 
\hline
Scheduler                  & LambdaLR                                                                     & Using warmup=4000                                                                                                        \\ 
\hline
Max. output length          & 50                                                                           & Maximum decoder output size                                                                                              \\ 
\hline
Gradient accumulation step & 32                                                                           & To get 320 sentences                                                                                                     \\ 
\hline
\end{tabular}
}
\caption{Main parameters used in training our primary submission DFKI-MLT.2.}
\label{t:protoc}
\end{table*}

\subsection{Loss and optimizer}
In our experiments, we use a generalized loss. The general loss is considering both vision and text as a single model so the backpropagation starts from the last layer of the language part to the first layer of the visual one.
We used the regular cross entropy loss from \citep{Vaswani:2017}, with smoothing value = 0.1. Our optimizer has the following configuration: Adam with beta values =(0.9, 0.98), epsilon =1e-8, weight decay = 0.001.

\section{Setup and Experiments}
\label{s:experiments}

\subsection{Data description}

\begin{table}
\centering
 \resizebox{\columnwidth}{!}{%
\begin{tabular}{|l|r|r|l|} 
\hline
\textbf{Corpus} & \textbf{Sentences} & \textbf{Vocab} & \textbf{Min/Mean/Max/Std}  \\ 
\hline
SRF+FN        & 17192                    & 26250               & 1/13.62/168/7.33             \\ 
\hline
SRF             & 7056                     & 14573               & 1/14.29/126/7.29             \\ 
\hline
FN              & 10136                    & 16723               & 1/13.15/168/7.32            \\
\hline
SRF+FN dev      & 420 & 2003 & 2/13.98/44/6.95 \\ 
\hline
\end{tabular}
}
\caption{Text corpus statistics in tokens.}
\label{tab:data}
\end{table}

\begin{table}
\centering
 \resizebox{\columnwidth}{!}{%
 \begin{tabular}{|l|r|r|r|r|r|} 
\hline
\textbf{Corpus}  & \textbf{Videos} & \textbf{Min} & \textbf{Mean} & \textbf{Max} & \textbf{Std}  \\ 
\hline
SRF  & 29              & 1492.6       & 1935.9       & 2106.2       & 106.3         \\ 
\hline
FN  & 197             & 209.8        & 349.3        & 571.4       & 64.1         \\
\hline
SRF+FN dev &         420   &   0.6 & 5.84  & 19.86 & 3.42\\ 
\hline
\end{tabular}
}
\caption{Number of videos and video statistics in seconds.}
\label{tab:data2}
\end{table}

For the submission, we use only the training and validation data given for the shared task and made up of FocusNews and SRF corpora, both parallel in Swiss German Sign Language and German text. SRF contains longer videos (approximately 30 minutes), FN contains more videos but shorter ones (approximately 5 minutes). The statistics of the German part of the corpus are summarized in Table~\ref{tab:data} and the video statistics in Table~\ref{tab:data2}. 


\subsection{Data preprocessing and batching}

Since the input videos are long and contain more than one sentence (Table~\ref{tab:data2}), we perform a subclipping step as preprocessing.
By reading the subtitle files entries (\emph{srt} in Figure~\ref{fig:archm}(a)), we extract the time intervals and the corresponding sentences. We use ffmpeg to cut videos using these timestamps. We save the resulting subclips and add paths with subtitles (sentences) in one single annotation file. 

We resize our input images to 224x224 pixels to leave a door open for pretraining approaches later. 
The batching is done using the Videodataset class \citep{wangtsn17}. The depth is the number of frames in a video, it constitutes the third dimension in the 3D model. 
In our experiments, we initialize it to 100. 
To make sure that the language model keeps its original performance, we need to simulate a higher batch size. However, only a small number of videos can be placed in the same batch. We use gradient accumulation and update every 320 sentences for this purpose. 

We do not do any preprocessing for the German textual data besides tokenization.

\subsection{Experimental protocol}
For the sake of reproducibility, we detail the setup for our primary submission in Table \ref{t:protoc}.

\subsection{Evaluation}
We use the same automatic metrics used by the shared task organisers in their preliminary automatic evaluation results~\cite{muller-etal-2022-findings}. We use SacreBLEU~\cite{post-2018-call} to calculate BLEU%
\footnote{BLEU|nrefs:1|bs:1000|seed:12345|case:mixed|eff:no| tok:13a|smooth:exp|version:2.2.0}
~\cite{papineni-etal-2002-bleu} and chrF2++%
\footnote{chrF2++|nrefs:1|bs:1000|seed:12345|case:mixed|eff:yes| nc:6|nw:2|space:no|version:2.2.0} \cite{popovic-2017-chrf}.
As semantic metric we use BLEURT%
\footnote{BLEURT v0.0.2 using checkpoint BLEURT-20}~\cite{sellam-etal-2020-bleurt}.

\begin{table*}[t]
\centering
\begin{tabular}{|l|r|r|r|} 
\hline
\textbf{VisualModel} & \textbf{BLEU}     & \textbf{chrF2++}      & \textbf{BLEURT}       \\ 
\hline
ResNet50\_3D         & 0.07 ± 0.02          & 8.07 ± 0.24          & 0.054 ± 0.003         \\ 
\hline
ResNet34\_3D         & \textbf{4.82 ± 0.99}          & 8.28 ± 0.60          & 0.075 ± 0.007         \\ 
\hline
ResNet10\_3D         & 2.83 ± 1.41          & \textbf{11.85 ± 1.32}         & \textbf{0.100 ± 0.012}           \\ 
\hline
\end{tabular}
\caption{Results from different 3D ResNet scales on the development set.}
\label{t:resultsDev}
\end{table*}

\begin{table*}[!t]
 \resizebox{\textwidth}{!}{%
\begin{tabular}{|l |rrr| rrr| rrr|}
\hline
\textbf{Submission} & \mc{3}{c|}{\bf BLEU} & \mc{3}{c|}{\bf chrF2++}& \mc{3}{c|}{\bf BLEURT} |\\
\hline
 & \mc{1}{c}{all} & \mc{1}{c}{SRF} & \mc{1}{c|}{FN} & \mc{1}{c}{all} & \mc{1}{c}{SRF} & \mc{1}{c|}{FN} & \mc{1}{c}{all} & \mc{1}{c}{SRF} & \mc{1}{c|}{FN}  \\
\hline
{\bf UZH (Baseline)}   & 0.12$\pm$0.06 & 0.09$\pm$0.03 & 0.19$\pm$0.11 &  4.7$\pm$0.4 &  4.5$\pm$0.5 &  5.0$\pm$0.7 & 0.102$\pm$0.006 & 0.095$\pm$0.006 & 0.110$\pm$0.009 \\
\hline
DFKI-MLT.1       & 0.07$\pm$0.05 & 0.05$\pm$0.02 & 0.12$\pm$0.10 &  6.2$\pm$0.4 &  5.9$\pm$0.5 &  6.4$\pm$0.5 & 0.100$\pm$0.008 & 0.097$\pm$0.009 & 0.100$\pm$0.012 \\
{\bf DFKI-MLT.2} & 0.11$\pm$0.06 & 0.08$\pm$0.03 & 0.17$\pm$0.13 &  6.3$\pm$0.4 &  6.4$\pm$0.6 &  6.1$\pm$0.6 & 0.083$\pm$0.008 & 0.074$\pm$0.008 & 0.091$\pm$0.013 \\
DFKI-MLT.3       & 0.08$\pm$0.04 & 0.06$\pm$0.02 & 0.13$\pm$0.10 &  6.1$\pm$0.4 &  6.3$\pm$0.6 &  6.0$\pm$0.6 & 0.075$\pm$0.009 & 0.067$\pm$0.009 & 0.081$\pm$0.014 \\
DFKI-MLT.4       & 0.02$\pm$0.01 & 0.02$\pm$0.01 & 0.04$\pm$0.02 &  3.9$\pm$0.2 &  3.7$\pm$0.3 &  4.1$\pm$0.3 & 0.066$\pm$0.004 & 0.063$\pm$0.004 & 0.070$\pm$0.008 \\
DFKI-MLT.5       & 0.04$\pm$0.02 & 0.03$\pm$0.00 & 0.08$\pm$0.04 &  5.2$\pm$0.2 &  4.9$\pm$0.3 &  5.5$\pm$0.4 & 0.078$\pm$0.004 & 0.074$\pm$0.005 & 0.080$\pm$0.007 \\
\hline
    \end{tabular}
}
 \caption{Automatic evaluation of our 5 submissions and the shared task baseline on WMT-SLT test set (all), the SRF subset and the Focus News (FN) subset as provided by the organizers~\cite{muller-etal-2022-findings}. DFKI-MLT.2 is our primary submission.} 
 \label{tab:automaticEval}
\end{table*}

\begin{table}[!t]
\centering
\resizebox{\columnwidth}{!}{%
\begin{tabular}{|l|l|} 
\hline
\textbf{Hypothesis} & \textbf{Reference}         \\ 
\hline
Die -.              & \begin{tabular}[c]{@{}l@{}}Die Diamantenschleiferei beschäftigt\\~63 Angestellte , davon 17 Behinderte , \\sowohl Rollstuhlfahrer als auch Gehörlose .\end{tabular}  \\ 
\hline
Der  - .            & \begin{tabular}[c]{@{}l@{}}Man arbeitet von 2004 bis 2009 \\ausbildungstechnisch mit dem \\Plussport Behindertensport Schweiz zusammen .\end{tabular}                \\ 
\hline
Und .               & \begin{tabular}[c]{@{}l@{}}3 . Für die Sommer Deaf Olympics 2017 \\standen mehrere Städte zur Auswahl , \\nämlich Barcelona , Buenos Aires und Ankara .\end{tabular}   \\
\hline
\end{tabular}
}\caption{Sample outputs in the translation of the development set by the DFKI-MLT.3 system.}
\label{t:tabSample}
\end{table}

\section{Results and Analysis}
\label{s:results}

Our best model according to BLEU is obtained with the largest 3D ResNet model and reaches 4.8 points on the development set, much higher than the performance of any system on the official test set. However, different metrics do not correlate, and chrF2++ and BLEURT ---which correlate better with human judgments than BLEU--- point towards a different model. Table \ref{t:resultsDev} shows how performance varies depending on the size of the 3D ResNet model. The  smallest models seem to perform better across metrics and therefore we use ResNet10\_3D in our submissions.

The low scores obtained with all our models correspond to a system that simply matches 
the most frequent words like "Die", "Der", "Und" as illustrated in Table \ref{t:tabSample}. The rest of the generated sentence is a series of <UNK> tokens that are removed after decoding. 
We observe that training passes through some remarkable steps. It starts to output the most frequent words repeatedly, 1-grams, and as training advances the system starts to predict higher $n$-grams. In our experiments, the model stayed at the 1-gram stage.

We submitted 5 runs to the shared task, three of them using ResNet10\_3D and the parameters are provided in Table~\ref{t:protoc}. DFKI-MLT.1 was created with our main system using a checkpoint before the end of the training, DFKI-MLT.2 is the best checkpoint. We realized that both submissions had encoding issues and contain <UNK> tokens.  We therefore sent a follow-up submission for DFKI-MLT.2, DFKI-MLT.3, containing the corrected format and without <UNK> tokens. As its translation quality was not even 0.5 BLEU points in the leaderboard, which may be less than a random walk from the vocabulary, we sent random walk results with repetitions (DFKI-MLT.4 and  DFKI-MLT.5) to compare the performance. 


A preliminary automatic evaluation has been made available by the organizers and it is shown in Table \ref{tab:automaticEval}. Our final submission reached $0.11\pm0.06$ BLEU, $6.3\pm0.4$ chrF2++ and $0.083\pm0.008$ BLEURT, where confidence intervals are at 95\% level. Results are therefore not statistically better than the baseline at 95\% level. Interestingly, according to BLEURT, the random walk though the vocabulary is not significantly worse than the combination of 3D CNNs and Transformers. 

In general, translation quality is always very bad, but results are slightly better for the FocusNew subset. FocusNews' input videos are shorter and this might imply a better alignment between videos and subtitles, improving the training. Some of our test outputs contain repetitions of (parts of) sentences from FocusNew dataset. Since this subcorpus is dominant in the final training (Table \ref{tab:data}) the system is biased towards its vocabulary and this also explains the better performance in its subtest.






\section{Conclusion}
\label{s:conclusions}

This paper presented an overview and some insights on spatio-temporal sign language representation which were used in the DFKI-MLT submission for the WMT-SLT 2022 shared task. To achieve our goal of building a fully end-to-end sign language model, we worked closely on the representation learning of visual features. Most of previous techniques for SLT simplify the feature representation by extracting spatial features and not benefiting from temporal features. This choice is motivated by the lack of temporal boundaries in sign language videos. To extract spatio-temporal features one can use 2D + 1D CNN approaches but this does not allow a fully end-to-end training as it still requires pretraining in another well-resourced task like object classification. In order to construct a specific representation model for SL and learn temporal modeling in a single model, we choose 3D CNNs and trained them from scratch simultaneously with the textual counterparts. 

The translations produced by this architecture are very short and output only high frequency tokens; in few cases, full fluent and grammatical sentences are constructed but their meaning unrelated to the source. The generation of short sentences might be a limitation of our approach that builds a sentence representation with an output conversion method that does not split a sentence in subunits that can be weighted by the Transformer's attention mechanism to generate the output.


However, all the systems in this shared task's leaderaboard have translation scores close to zero. 
This shows the extreme difficulty of SLT and how bad current systems generalize to new data sets. We believe that system comparisons with such a bad translation quality do not allow to extract meaningful conclusions. In our future work, we investigate on different temporal modeling coupled with the 3D CNNs approach to further pursue the goal of developing a high-quality end-to-end system.



\bibliography{anthology,custom}
\bibliographystyle{acl_natbib}

\end{document}